\documentclass[10pt,twocolumn,letterpaper]{article}

\usepackage{iccv}
\usepackage{times}
\usepackage{epsfig}
\usepackage{graphicx}
\usepackage{amsmath}
\usepackage{amssymb}

\usepackage[pagebackref=true,breaklinks=true,letterpaper=true,colorlinks,bookmarks=false]{hyperref}

\iccvfinalcopy 

\def\iccvPaperID{****} 

\begin{document}

\title{DuBox: No-Prior Box  Objection  Detection via Residual Dual Scale Detectors\vspace{-5mm}}
\author{
 Shuai Chen \quad  Jinpeng Li \quad Chuanqi Yao \quad Wenbo Hou \quad Shuo Qin \quad Wenyao Jin \quad Xu Tang\\
 Baidu Inc.\\
 {\tt\small chenshuai11,lijinpeng,yaochuanqi,houwenbo,qinshuo01,jinwenyao,tangxu02@baidu.com}
}
\maketitle

\begin{abstract}\vspace{-1mm}
Traditional neural objection detection methods use multi-scale features that allow multiple detectors to perform detecting tasks independently and in parallel. At the same time, with the handling of the prior box, the algorithm's ability to deal with scale invariance is enhanced. However, too many prior boxes and independent detectors will increase the computational redundancy of the detection algorithm. In this study, we introduce Dubox, a new one-stage approach that detects the objects without prior box. Working with multi-scale features, the designed dual scale residual unit makes dual scale detectors no longer run independently. The second scale detector learns the residual of the first. Dubox has enhanced the capacity of heuristic-guided that can further enable the first scale detector to maximize the detection of small targets and the second to detect objects that cannot be identified by the first one. Besides, for each scale detector, with the new classification-regression progressive strapped loss makes our process not based on prior boxes. Integrating these strategies, our detection algorithm has achieved excellent performance in terms of speed and accuracy. Extensive experiments on the VOC, COCO object detection benchmark have confirmed the effectiveness of this algorithm.
\end{abstract}\vspace{-5mm}



\section{Introduction}\label{sec:introduction}

Object detection has been a challenging issue in the field of computer vision for a long time. With the development of deep neural networks (DNN), significant progress has been made in object detection in recent years. It is a prerequisite for a variety of industrial applications, such as autonomous driving \cite{liang2018cirl} and face analysis\cite{tang2018pyramidbox}. Due to the advancement of deep convolutional neural networks \cite{vgg16,residualnet} and well-annotated data sets\cite{voc,mscoco}, the performance of object detectors has been significantly improved.

\begin{figure}[h]
	\centering  
	\includegraphics[width=1\linewidth]{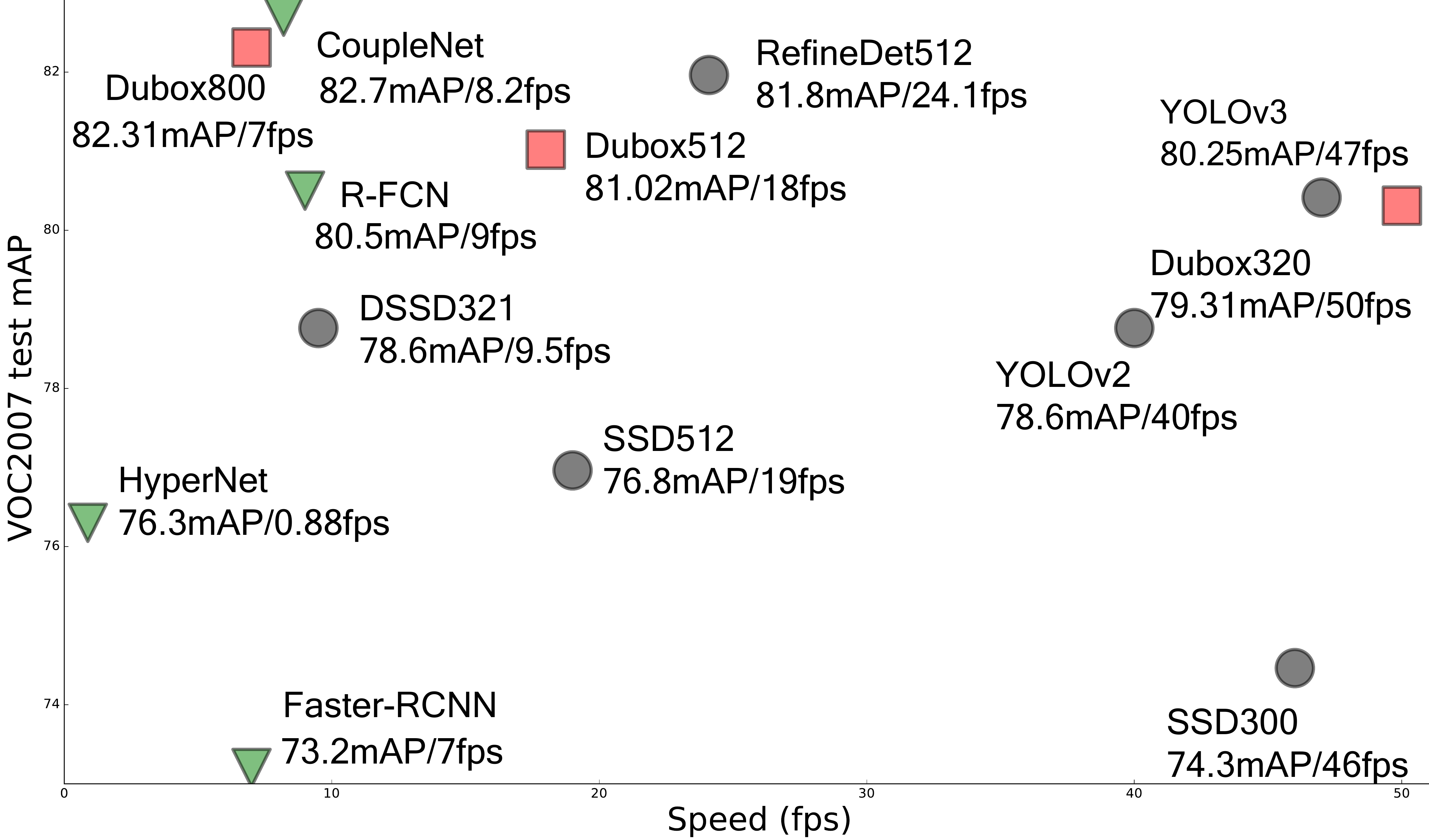}
	\caption{Some comparisons with the precision and speed to classical algorithms on VOC07, $\bigtriangledown$ is two-stage method, $\bigcirc$ denote the one-stage algorithm, $\Box$ is Dubox.}  
	\label{fig:speed}  
\end{figure}
Images in the real world contain different scale objects. Scale variation has become a challenging problem in the field of objection detection. To achieve scale invariance, state-of-the-art approaches typically combine features of multiple levels to construct a feature pyramid or multi-layer feature tower. Meanwhile, to improve the detection performance, the multi-scale method uses multiple detectors in parallel at various scales. For example, RetinaNet\cite{retinanet} has five scale detectors ($p3$-$p7$) that are detected in parallel on the feature pyramid structure\cite{fpn}. YOLOv3\cite{yolov3coco} has three detectors running on the main network.

In addition, the prior box is considered to be an effective means for dealing with scale invariance. It is fundamental for lots of detectors, e.g., anchors in Faster RCNN\cite{fasterrcnn} and YOLOv2\cite{yolov2}, default boxes in SSD\cite{ssd}. Prior boxes are a bunch of boxes with pre-defined sizes and aspect ratios that tile the feature map in a sliding window manner, to serve as detection candidates. 
The prior box discretizes the space of possible output bounding-box shapes, and DNN regresses bounding-boxes based on a specific prior-box taking advantage of prior information. Hybridising of multi-scale detection and prior boxes is a common practice in state-of-the-art detectors, which takes advantage of multi-scale features and pre-computed bounding box statistics.

In multi-scale detectors, a specific feature level is responsible for objects with similar scales. The correspondence between the object scale and the feature level is found independently of each other by heuristic-guided feature selection and different detectors. However, one of the drawbacks of this design is computation redundancy. Spatial scales of features are discrete, which violate the scale continuity in the real world. Each scale detector will try to detect objects lying in the intermediate scales of them. To regress multiple scale objects, more detectors are needed in this design. Meanwhile, a large number of anchor boxes are required to overlap with object bounding boxes sufficiently. For example, there are more than 100k anchors in RetinaNet\cite{retinanet}, which results in an imbalance between positive and negative samples. These strategies increased the computational complexity. Although, in early stage, some works have explored the no-prior box detection like YOLOv1 \cite{yolov1} and Densebox\cite{densebox}, they achieved high speed in a sacrifice of accuracy.  Recently, some algorithms such as CornerNet\cite{cornernet} designed the paired key points to remove the anchor settings and delivered excellent performance.
 
Based on these observations, we explore two issues in this paper$:$
Is it possible to achieve excellent performance using less scale detectors?
Is it possible to regress accurate bounding boxes without anchor?
In other words, can we design high-performance dual scale detectors algorithms that doesn't use the prior box?

In this paper, we introduce Dubox method, an one-stage approach for object detection without prior boxes. Dubox classifies objects and regresses bounding boxes directly in one network. The object detection problem becomes a pixel-wise classification-regression problem. To alleviate the influence of scales, we also design a dual-scale structure called dual scale residual unit, which allows dual scale detectors to no longer run independently. In Dubox, the second scale detector learns the residual of the first. In this residual design, we add some computational redundancy reduction strategies by enhancing the capability of heuristic-guided. This method can further encourage the first scale detector to maximize the detection ability of small targets, and let the second scale detector to detect objects that cannot be identified by the first one. Although the two independent scales do not have good performance when test separately, the joint inference results have achieved outstanding performance. At the same time, for each detector, our detection framework with the new classification-regression progressive strapped loss makes our process not based on prior boxes. Integrating these strategies, our detection algorithm has achieved excellent performance in terms of speed and accuracy. Extensive experiments on the VOC\cite{voc}, COCO\cite{mscoco}object detection benchmark confirm the effectiveness of our method. Some comparisons with the precision and speed to classical algorithms are shown in Fig.\ref{fig:speed}.

\section{Related Work}\label{sec:related_work}
\textbf{Detection by single-scale}

Single-scale detectors detect targets in a typical ratio and cannot identify objects in other proportions. To overcome this drawback, many algorithms use image pyramids, and each proportional object in the pyramid is fed into the detector. Such framework design is prevalent in algorithms that do not use deep learning, and often design some manual features such as HOG \cite{hog} or SIFT\cite{sift}. There are also algorithms using this method in the CNN network, like \cite{li2015convolutional}. The detector only processes the features in a specific range, and based on this feature map, classifies and regresses the object box. Although this may reduce the detection difficulty of the detector, on the whole, it has a substantial computational cost, making it not easy to use on devices with low computing capability, which greatly limits its practicality.

\textbf{Detection by multi-scale}

The multi-scale detection algorithm only needs a fixed-scale input and detects objects with diverse sizes. YOLOv3\cite{yolov3coco} and RetinaNet\cite{retinanet} had fixed input sizes and detected in parallel at various scales by using multiple detectors. In general, detectors in the lower layers detect small targets and the uppers are more accessible to identify large objects. This is a heuristic-guided strategy. The addition of the anchor design further strengthens this guidance. However, because multiple levels of detectors operate independently in parallel in each scale feature, there is no cooperation between them, resulting in a large amount of detection redundancy. In the meanwhile, the common design of anchor in these detectors dramatically increases the number of output channels and aggravates the computation burden.


\section{Our Method}\label{sec:our_detector}

In this section, we depict each component of our pipeline (Fig.\ref{fig:residualdualshot}) in detail. We first devise the classification and regression target label maps for our no-prior box detector in Section \ref{sec:no_prior_box_Detection}.

The dual scale residual unit is designed for letting the high-level detector to learn the residual of the low-level one, which is described in Section \ref{sec:residual_dual_shot}. To get ride of prior boxes and make the detector's classification and regression work synergistically, we designed classification-regression progressive strapped loss, which  will be explained in Section \ref{sec:crp_loss}. Many redundancy reduction strategies are added to enhance the capability of heuristic guiding in \ref{sec:mrs}. In Section \ref{sec:dataaugment}, we represent the positive and negative sample balance and data augmentation strategies for the detector during the training phase.

\subsection{No-prior Box Detection}\label{sec:no_prior_box_Detection}
In this section, we describe the method to generate targets for classification and regression. We transform bounding box ground truth represented by coordinates to pixel-wise label maps.

\textbf{Hooks}

Dubox is a single neural network unifying all necessary components of object detection. The detector design enables end-to-end training and real-time inference while maintaining high average precision.

\begin{figure}[h]
	\centering  
	\includegraphics[width=1\linewidth]{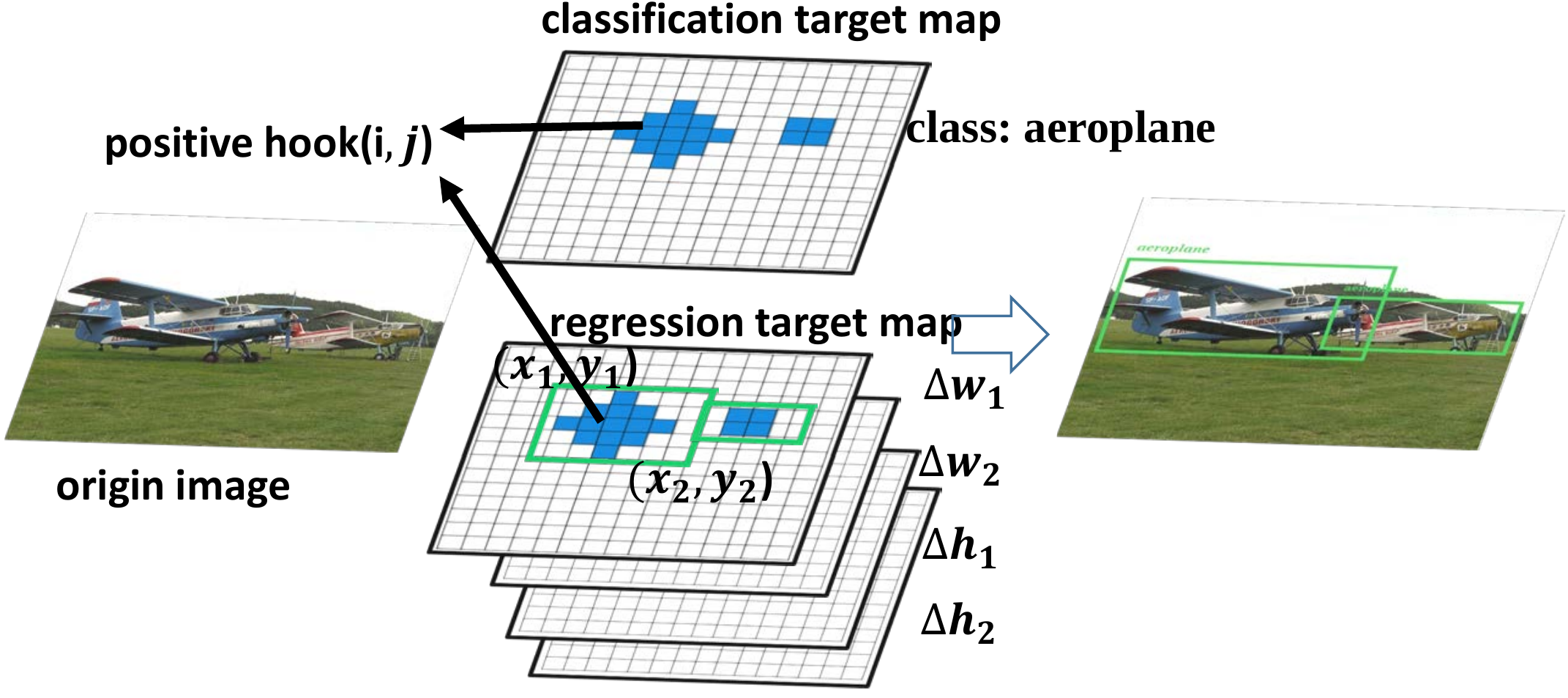}
	\caption{DuBox use fixed hook$(i,j)$ to unite bounding box prediction and classification. The blue points are positive hooks, others are negatives.}  
	\label{fig:label}  
\end{figure}

Our network takes the whole image as input and predicts the result feature maps with the down-sampling level of $s$-times. Supposing the output map size is $(h,w)$, we define the location $(i,j)$ in output as hook, where $i\in \left[ \text{0,}w \right)$ and $j\in \left[ \text{0,}h \right)$. Dubox predicts each bounding box and its confidence scores of all categories at each hook on the output feature, as shown in Fig.\ref{fig:label}.

Note that hooks are parameters predefined by the network output. They represent the positions of each points on the output map. We will use this feature to design the target maps for classification and regression. 

\textbf{Classification and regression target map}

 Suppose there are $w\times h$ hooks in output. An bounding box $(x_{1},y_{1},x_{2},y_{2})$ of an object in the output map represents its left-top and right-down corner points. They are sample mapping from location $(x_{1}s,y_{1}s,x_{2}s,y_{2}s)$ in origin image by stride $s$.  we define the positive range $\varTheta$ with the following condition:

\begin{equation}\label{eq:range}
\left( i-\left( x_2+x_1 \right) /2 \right) ^2+\left( j-\left( y_2+y_1 \right) /2 \right) ^2\leqslant r^2,
\end{equation}
where $r=\sqrt{\left( x_2-x_1 \right) ^2+\left( y_2-y_1 \right) ^2}\,/p$.   That means if a hook $(i, j)$ falls into the range $\varTheta$ of an bounding box, then it's responsible for detecting the corresponding object. Each hook predicts one bounding box$\left( Pr_{\varDelta w_1},Pr_{\varDelta w_2},Pr_{\varDelta h_1},Pr_{\varDelta h_2} \right)$ and one confidence score $Pr_{cls}$ for this object. This confidence score reflects how confident the model is that the hook is in the range of the object and also how accurate it thinks it is belong to an class is that it predicts. $p$ is an predefined value for adjusting the range. The size of this value will affect the number and proportion of hooks that large objects and small objects occupy in detecting.  We will discuss it further in Section \ref{sec:mrs}.

For regression target, traditional methods\cite{fasterrcnn,ssd} regress the center$(c_{x},c_{y})$, width $w$ and height $h$. Each bounding box consist of 4 predictions: $(c_{x},c_{y},w,h)$. However, with the position $(i,j)$, because the regression box can adjust all of its offset values, but the locations of classification cannot change. This approach will result in inconsistency between classification and regression.

As shown in Fig.\ref{fig:label}, we designed a hook-based regression target that each bounding box consists of $4$ predictions: $\left(\varDelta w_1,\varDelta w_2,\varDelta h_1,\varDelta h_2 \right)$ which represent the offset to the positive hooks $(i,j)$ in an object: 
\begin{equation}\label{eq:bbox_label}
\begin{array}{c}
	\varDelta w_1=i-x_1,\varDelta w_2=x_2-i\\
	\varDelta h_1=j-y_1,\varDelta h_2=y_2-j.\\
\end{array}
\end{equation}

With such design, each hook center must reside in its own box prediction. Consequently, the result of the classification and the result of the regression will not have inconsistency by predicting different objects in the image. In inference phase, using the fixed hook $(i,j)$, and predict offset $\left( Pr_{\varDelta w_1},Pr_{\varDelta w_2},Pr_{\varDelta h_1},Pr_{\varDelta h_2} \right)$, we can obtain the bbox results in origin image by:
\begin{equation}
\label{eq:get_bbox}
\begin{array}{c}
	x_1=\left( i-Pr_{\varDelta w_1} \right) s,x_2=\left( i+Pr_{\varDelta w_2} \right) s\\
	y_1=\left( j-Pr_{\varDelta h_1} \right) s,y_2=\left( j+Pr_{\varDelta h_2} \right) s.\\
\end{array}
\end{equation}

In our Dubox detector, we use two different down-sampling scales $detector_{1}$ $s =8$, $detector_{2}$ $s=32$. Thus, with the input  $(w_{ori},h_{ori})$ image, our final prediction is a $\left( \frac{w_{ori}}{8}\times \frac{h_{ori}}{8} \right) \times C\times 5$ tensor in $detector_{1}$, $\left( \frac{w_{ori}}{32}\times \frac{h_{ori}}{32} \right) \times C\times 5$ tensor in $detector_{2}$, where $C$ is the number of class. For PASCAL VOC dataset $C = 20$ and for COCO dataset $C = 80$.

\subsection{ Residual Dual Scale Detectors}\label{sec:residual_dual_shot}
 \begin{figure*}
	\centering  
	\includegraphics[width=1\linewidth]{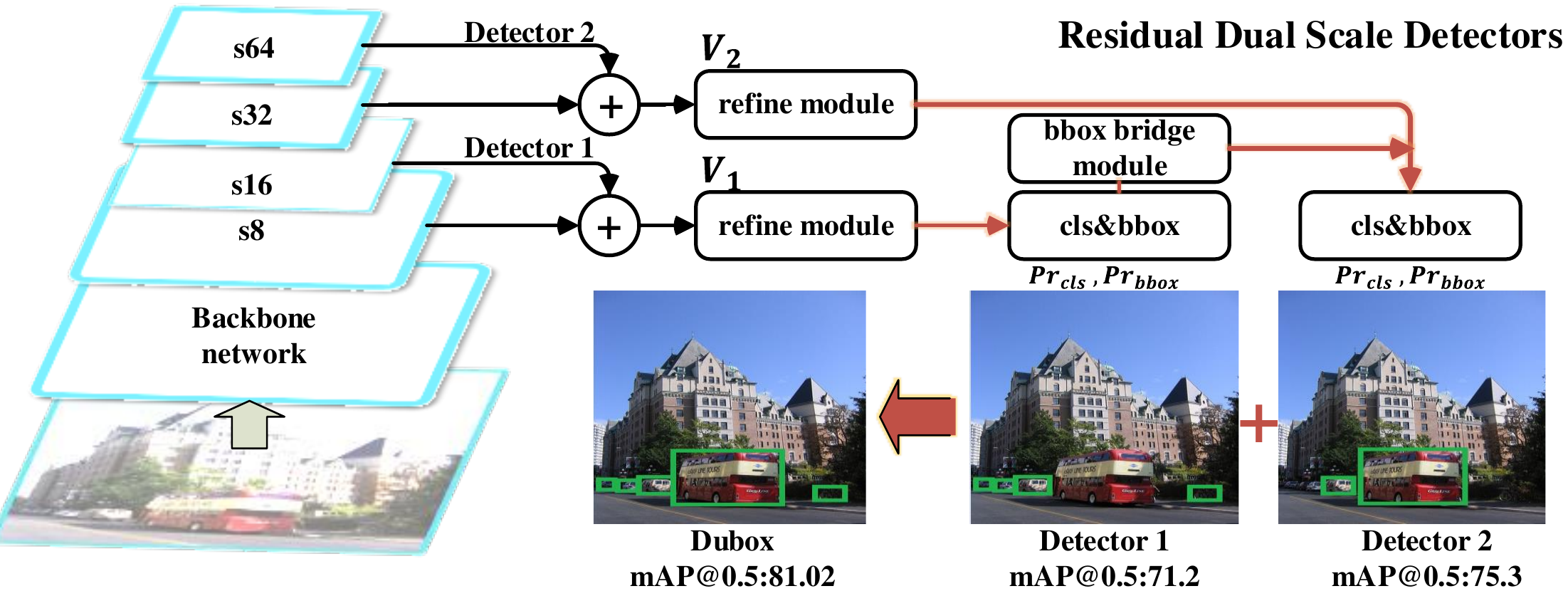}
	\caption{The residual dual scale detectors structure, the $detector_{2}$ will learn the residual of $detector_{1}$ through an down-sample bbox bridge module. }  
	\label{fig:residualdualshot}  
\end{figure*}

 Dual scale residual unit is a sub-structure based on a shared feature extraction backbone. The residual dual scale detector combines the features of different levels detectors by sharing the feature extraction network such as VGG-16\cite{vgg16}, ResNet\cite{residualnet}. The structure of residual unit contains two detectors where the high-level detector will learn the residual of regression boxes found in low-level detector. The detailed structure is shown in Fig. \ref{fig:residualdualshot}. $Detector_{1}$ connects the features at $s$=8 and $s$=16, $detector_{2}$ adds features at $s$=32 and $s$=64. De-convolution (stride 2, $1\times1$ kernel and 256 channels) is used in up-sampling feature maps of different scales to the same spatial size. We do not connect the features of $s$ = 32 and $s$ = 16 used in FPN\cite{fpn} structure.

\textbf{Refine module}

In our network, an object box in output map contains positive and negative hooks, which requires our system to consider the surrounding situation when classifying the hooks.
Combining the different scales, hooks can learn features with enlarged receptive fields. After mixing feature maps, every scale are connected to a refine module, which is a simple implementation of channel and spatial attention model \cite{attention}:
\begin{equation}
\begin{aligned}
	\boldsymbol{\gamma }&=\Phi \left( \boldsymbol{V} \right) ,\\
	\boldsymbol{x}&=f\left( \boldsymbol{V,\gamma } \right),\\
\end{aligned}
\end{equation}
 where $\textbf{\textit{V}}$ is the input feature of the refine module. $f()$  is a multiplication for feature map regions and corresponding region weights, $\Phi$ is refine module. 
 
\begin{figure}
	\centering  
	\includegraphics[width=1\linewidth]{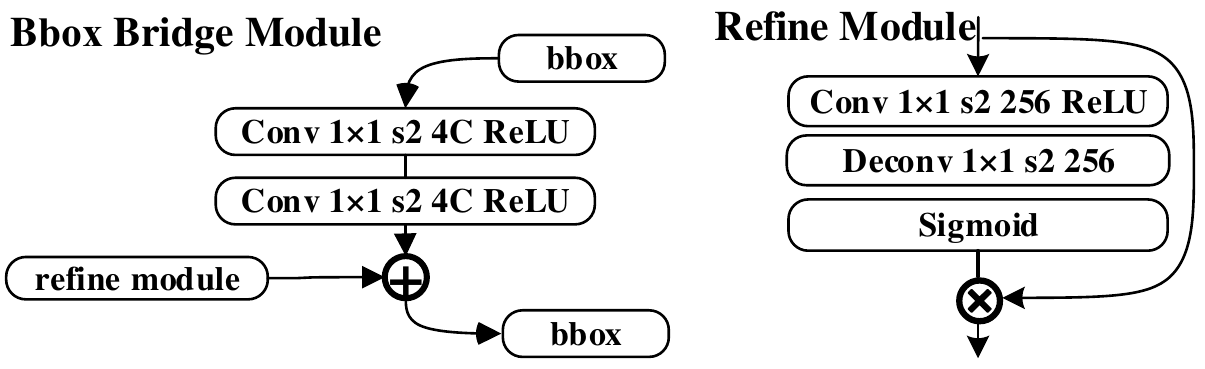}
	\caption{The detail structure of bbox bridge module and refine module. }
	\label{fig:modules}  
\end{figure}

The detail design of refine module is shown in Fig.\ref{fig:modules}, where $Sigmoid\left( x \right) =\frac{1}{1+e^{-x}}$ and $\text{Re}LU\left( x \right)=\max \left( \text{0,}x \right)$. In the structure, we use an convolution (stride 2, $1\times1$ kernel and 256 channels) and De-convolution (stride 2, $1\times1$ kernel and 256 channels) to reduced by 2 times and enlarge the feature map back to the input size of the refine module, this technique helps our detector to push further the ability of considering the around feature for prediction.

\textbf{Bbox bridge module}

Bbox (bounding box) bridge module connects the regressions of low-level and high-level detector, so that the high-level regression is based on the low-level residuals. We inductively describe residual dual scale detectors as follows:
\begin{equation}
\begin{aligned}
	&Pr_{bbox}^{b+1}\left( \boldsymbol{V}_{\boldsymbol{b}+\boldsymbol{1}} \right) =\phi \left( Pr_{bbox}^{b}\left( \boldsymbol{V}_{\boldsymbol{b}} \right) \right) +\tau \left( \boldsymbol{V}_{\boldsymbol{b}}\, \right)\\
	&Pr_{bbox}^{1}\left( \boldsymbol{V}_{\boldsymbol{1}} \right) =\tau \left( \boldsymbol{V}_{\boldsymbol{1}}\, \right),\\
\end{aligned}
\end{equation}
where $\boldsymbol{V}_{\boldsymbol{b}}$ is the input feature map  of $detector_{b}$, $Pr_{bbox}\left(\boldsymbol{V}_{\boldsymbol{b}} \right)$ denote $detector_{b}$ predict the bbox with the input $\boldsymbol{V}_{\boldsymbol{b}}$ and it equal to $\tau(\boldsymbol{V}_{\boldsymbol{b}})$ when $b = 1$ , $\phi( )$ is the bbox bridge module, it contains two convolution (stride 2, $1\times1$ kernel,$4c$ channels). The bbox bridge module transmits the residual of low-level to high one by stride 4. The details structure is shown in Fig.\ref{fig:modules}.

Consequently, the residual dual scale detectors make the $detector_{2}$ to perform residual learning based on the prediction of the $detector_{1}$. This method makes multi-detector in our design not independent as the higher scale depends on the low-level's results. 

\subsection{Classification-Regression Progressive Strapped Loss}\label{sec:crp_loss}

In the anchor-based method, with the help of the prior box, the detector has prior knowledge of box shapes. And it performs its prediction by adjusting the pre-defined anchor shape which boosts their fitting ability of around anchors. Dubox donot have any prior box shape, we have to design a more robust classification and regress strategy, primarily the loss function.  

\begin{figure}[h]
	\centering  
	\includegraphics[width=1\linewidth]{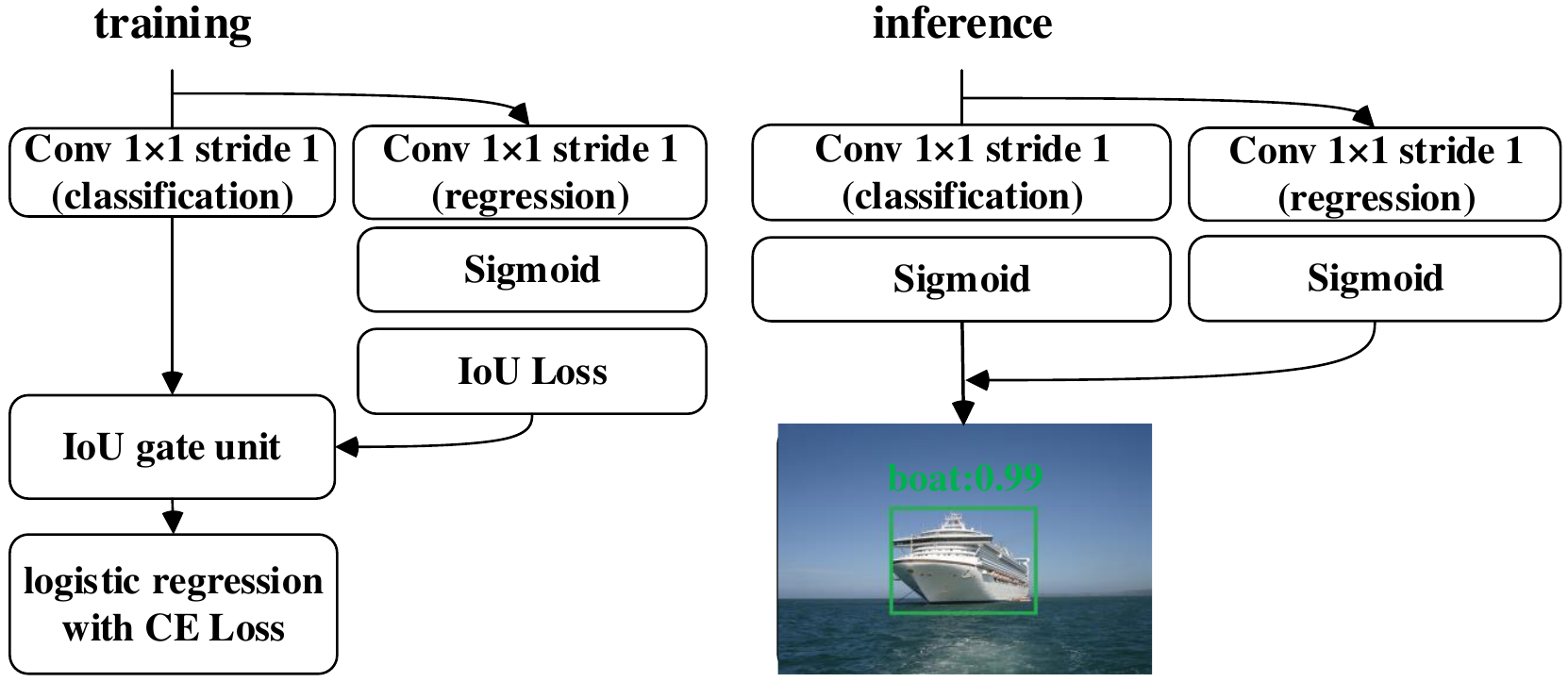}
	\caption{In the training phase, classification-regression progressive strapped loss(CRPS loss) working with the IoU gate unit. With the classification and regression hooks, we can get the detecting results by Eq.\ref{eq:get_bbox} in inference phase. }  
	\label{fig:cr_loss}  
\end{figure}

In order to regress the bounding box target of the offset $\left( \varDelta w_1,\varDelta w_2,\varDelta h_1,\varDelta h_2 \right)$ to positive hooks without prior, a loss function which is robust to objects of varied shapes and scales is in need. IoU loss normalize the loss of boxes with different scales by their areas and show robustness to objects of various shapes and scales\cite{iouloss}. The mathematical form of IoU loss can be expressed as:
\begin{equation}
L_{bbox}=-\sum_{i,j\in \boldsymbol{\varTheta }}{ln\left( IoU\left( \boldsymbol{Pr}_{\boldsymbol{bbox}}^{\boldsymbol{i,j}},\boldsymbol{Gt}_{\boldsymbol{bbox}}^{\boldsymbol{i,j}} \right) \right)},
\end{equation}
where $\boldsymbol{Gt}_{\boldsymbol{bbox}}^{\boldsymbol{i,j}}$ is the ground truth  box of hook $(i,j)$, $\boldsymbol{Pr}_{\boldsymbol{bbox}}^{\boldsymbol{i,j}}$ denotes the the predict bbox in hook$(i,j)$. $IoU\left( \boldsymbol{Gt}_{\boldsymbol{bbox}}^{\boldsymbol{i,j}},\boldsymbol{Pr}_{\boldsymbol{bbox}}^{\boldsymbol{i,j}} \right)$ denotes the Intersection-over-union (IoU) between the predicted bounding box and ground truth.  For regression, we only regress to the positive samples and ignore the negative ones. In the actual implementation and show in Fig.\ref{fig:cr_loss}, we use Sigmoid to normalize the prediction to $[0,1]$. Correspondingly, we also map our predicted targets to $[0,1]$ and Eq.\ref{eq:bbox_label} change into：
\begin{equation}
\begin{array}{c}
	\varDelta w_1=\left( i-x_1 \right) /w,\varDelta w_2=\left( x_2-i \right) /w\\
	\varDelta h_1=\left( j-y_1 \right) /h,\varDelta h_2=\left( y_2-j \right) /h.\\
\end{array}
\end{equation}
For classification problems, logistic regression with cross entropy loss is widely used in objection detection methods. The classification loss function can be described as:
\begin{small} 
\begin{equation}
L_{cls}\,\,=-\sum_{i,j=0}^{h,w}{CE\left( \boldsymbol{Pr}_{\boldsymbol{cls}}^{\boldsymbol{i,j}},\boldsymbol{T}_{\boldsymbol{cls}}^{\boldsymbol{i,j}} \right)},
\end{equation}
\end{small} where $\boldsymbol{T}_{\boldsymbol{cls}}^{\boldsymbol{i,j}}$, $\boldsymbol{Pr}_{\boldsymbol{cls}}^{\boldsymbol{i,j}}$ is the class label and predict result of hook $(i,j)$, $CE()$ is the cross entropy function.

However, this classification and regression is flawed as the two losses are independent which results in inconsistency during prediction. Our experiments shows that the detector often predicts a right bounding box which the classification fails to predict the right class.

Based on this observation we rebuild the classification loss progressive strap by IoU:
\begin{small} 
\begin{equation}
\begin{aligned}
	L_{cls}&=-\sum_{i,j\in \boldsymbol{\varTheta }}{\begin{array}{c}
	CE\left( \boldsymbol{Pr}_{\boldsymbol{cls}}^{\boldsymbol{i,j}},\boldsymbol{T}_{\boldsymbol{cls}}^{\boldsymbol{i,j}} \right) \sigma \left( \boldsymbol{Pr}_{\boldsymbol{bbox}}^{\boldsymbol{i,j}},\boldsymbol{Gt}_{\boldsymbol{bbox}}^{\boldsymbol{i,j}} \right)\\
\end{array}}\\
	&-\sum_{i,j\notin \boldsymbol{\varTheta }}{CE\left( \boldsymbol{Pr}_{\boldsymbol{cls}}^{\boldsymbol{i,j}},\boldsymbol{T}_{\boldsymbol{cls}}^{\boldsymbol{i,j}} \right)},\\
\end{aligned}
\end{equation}
\end{small}
where $\sigma()$ is the IoU gate unit, it can be defined  as:
\begin{small} 
\begin{equation}
\sigma \left( \boldsymbol{Pr}_{\boldsymbol{bbox}}^{\boldsymbol{i,j}},\boldsymbol{Gt}_{\boldsymbol{bbox}}^{\boldsymbol{i,j}} \right) =\begin{cases}
	1&		if\,\,IoU\left( \boldsymbol{Pr}_{\boldsymbol{bbox}}^{\boldsymbol{i,j}},\boldsymbol{Gt}_{\boldsymbol{bbox}}^{\boldsymbol{i,j}} \right) >\epsilon\\
	0&		elsewise.\\
\end{cases}
\end{equation}
\end{small}

As shown in Fig.\ref{fig:cr_loss}, for positive hooks $(i,j)$, the classification includes it as a positive sample, only if the predicted and ground truth of the regression match $\epsilon$ overlaps. Otherwise, it's ignored. In our experiment $\epsilon$ is 0.5.

With the dual scale detectors, the final loss funciton:
\begin{equation}
L=\sum_{b=1}^2{\left( \lambda _{bbox}L_{bbox}+\lambda _{cls}L_{cls} \right),}
\end{equation}
where $\lambda _{bbox}$, $\lambda _{cls}$is the hyper-parameter used to keep the task of classification and regression of $detector_{b}$ in balance.

\subsection{ Reducing Redundancy Strategy}\label{sec:mrs}
The primary goal of the residual dual scale detectors is to maximize the overall fitting capacity of a multi-detector. In general, high-level detector are better at detecting large objects, while low-levels are more sensitive to small ones in image. To enhance this capability of heuristic guiding and reduce redundancy we adopt the following strategies:

\textbf{Differentiate positive range}

As mentioned in Eq.\ref{eq:range}, $p$ is a predefined value for adjusting the positive range.  The size of this value will affect the number and proportion of hooks between large and small objects occupy in detecting. In order to control the proportion of large and small samples in $detector_1$, we design $p$ is 10 in $detector_1$ and 9 in $detector_2$. At the same time, add a constraint to the positive range of $detector_1$ that
\begin{equation} r=\,\,arg\min \left( r,3 \right).\end{equation}
 This method ensures the numbers of large object positive hooks have a limit and this method improves the performance of low-levels in detecting small objects.

\textbf{Differentiate scale weight}

In order to further differentiate the detection capabilities of the two detectors. If the target bounding box of an object occupies an area greater than 0.3 in the original image, the regression of our $detector_1$ will ignore this object, and it can be described as：
\begin{equation}
L_{detector_1}=\sum_{i,j\in \boldsymbol{\varTheta }}{\boldsymbol{\lambda }_{\boldsymbol{bbox}}^{\boldsymbol{i,j}}L_{bbox}^{i,j}+\lambda _{cls}L_{cls}}
\end{equation}
where $\boldsymbol{\lambda }_{\boldsymbol{bbox}}^{\boldsymbol{i,j}}$ is zero if the target bbox of an object occupies an area greater than 0.3 in the original image, other situations are 1.  

\subsection{ Data Augment and Sample Balance}\label{sec:dataaugment}
We use several data augmentation strategies presented in \cite{ssd} to construct a robust model for adapting the variations of objects. That is, we randomly expand and crop the original training images with additional random photo-metric distortion and flipping to generate the training samples\cite{refinedet}. In additional to these methods, we use a batch balance method. 

\textbf{Batch balance}

We first traverse the entire data set to create a hash category table. For each image, we select the category of an object as the primary attribute in turn, which means that if a picture has $n$ objects, this picture will appear $n$ times in the hash table. In training, we select a category picture from the header of the hash category table with equal probability to fill the current batch in training phase.

\textbf{Positive and negative sample balance}

To mitigate the imbalance issue, we design a positive and negative sample balance and online hard example mining (OHEM)\cite{ohem}. Concretely, in the training phase, assuming the number of the positive hooks in $\varTheta$ is $N$, we will select $3N$ negatives in the output map through sorting the negative hooks by the loss and select the top-$3N$ negatives. The rest of negatives are ignored. This strategy is only used in classification task.

\section{Experiment}\label{sec:experiment}

\subsection{Implementation details}
As a common practice, the backbone network of our Dubox method is initialized by the pretrained VGG-16\cite{vgg16} and ResNet-101\cite{residualnet} classifier trained from Imagenet\cite{imagenet}, and the extra added convolution layers are randomly initialized by the $Xavier$ \cite{xavier} method. During training, all parameters of the network is fine-tuned on the detection datasets. For simplicity, all architectures including the dual scale residual unit of our Dubox are trained in the end-to-end manner. We train the network on 8 Nvidia P40 GPUs using the synchronized SGD with gradient clipping value 10. The momentum is 0.9 and weight decay is 0.0005.

\subsection{Ablation Experiments}
\subsubsection{Baseline}
We experiment with several variants of Dubox to demonstrate how the key components of it affect the detection performance. VOC 2007 and VOC 2012 trainval datasets are used to train the models, and all hyper training parameters and input size ($512\times512$) keep same among all models for fair comparison. VOC 2007 test set is used as the testing data. Smooth $l1$ loss\cite{fasterrcnn} is widely used as the box regression method by the anchor-based one stage and two stage detection methods and we use the loss without anchors as our baseline. Because the anchor-based methods have the prior box shape and size.  But no-prior box method don't have the prior knowledge. With the help of prior bix, the detector only get 54.3\%mAP.

In ablation experiments, we use the IoU loss with batch balance and OHEM as the box regression loss, and Tab.\ref{table:ablationexpriment} shows that Dubox (73.4\%mAP) can improve \textbf{9.84\%} mAP using the IoU loss compare to the smooth $l1$ loss with batch balance and OHEM (63.56\%mAP). 

\begin{table}[h]  
\centering  
\caption{Comparison of Dubox with different key components on pascal VOC 2007 test $(512\times512)$.}  
  \resizebox{0.5\textwidth}{!}{ 
\begin{tabular}{|c|cccccccc|}
\hline
Component               &     &            &           &           &    Dubox &          &          &       \\
\hline
Batch balance and OHEM? &     & \checkmark &\checkmark &\checkmark &\checkmark&\checkmark&\checkmark&\checkmark\\
IoU loss?               &     &            & \checkmark& \checkmark&\checkmark& \checkmark         &          &           \\
Hooks?               &     &            &           & \checkmark&\checkmark&\checkmark&\checkmark&\checkmark\\
Refine module?          &     &            &           &           &\checkmark&\checkmark&\checkmark&\checkmark\\
Residual dual scale?     &     &            &           &           &          &\checkmark&\checkmark&\checkmark\\
CRPS loss?              &     &            &           &           &          &          &\checkmark&\checkmark\\
Reducing redundancy strategy?      &     &            &           &           &          &          &          &\checkmark\\
\hline
mAP@0.5                &54.32&63.56       &73.4       &74.2       &75.7      &  77.01   &  79.43   &81.02\\
\hline
\end{tabular}
}
\label{table:ablationexpriment}  
\end{table} 

 \begin{table*}[h]
    \centering
    \caption{The performance of dual branch on VOC2007 dataset $(512\times512)$.}
     \resizebox{\textwidth}{!}{ 
    \begin{tabular}{|c|c|cccccccccccccccccccc|}
        \hline
        detector & mAP & aero & bike & bird & boat & bottle& bus& car & cat & chair& cow & table & dog & horse & mbike & person & plant & sheep & sofa & train & tv\\
        \hline
        $Detector_{1}$ & 71.26& 77.52 & 80.30 & 60.94 & 59.20 & 77.32& 81.50& 83.48 & 78.82 & 41.78& 59.35 & 69.24 & 69.39 & 80.61 & 78.66 & 71.48 & 32.89 & 60.95 & 60.78 & 67.01 & 68.70\\
           \hline
        $Detector_{2}$ & 75.37& 79.12 & 86.33 & 71.46 & 67.76 & 64.10& 82.91& 86.62 & 80.53 & 61.04& 80.62 & 72.77 & 79.90 & 84.60 &80.92 & 78.75 & 48.93 & 74.43 & 73.91 & 78.41 & 74.37\\
           \hline
         $Joint$ & 81.02 & 85.28 & 88.39 & 79.89 & 74.31 & 67.88& 86.71& 85.58 & 89.35 & 60.62& 86.78 & 73.37 & 88.82 & 88.62 & 85.03 & 83.21 & 55.43 & 84.42 & 82.89 & 86.02 & 77.85\\
         \hline
    \end{tabular}
    }
    \label{tab:dualshotperformace}
\end{table*}
\begin{table*}[h]
    \centering
    \caption{Comparison with state-of-the-art detectors on VOC 2007 and 2012}
    \begin{tabular}{|c|c|c|c|cc|}
        \hline
         Methods & Backbone & Input Size & FPS & mAP@0.5(VOC07) & mAP@0.5(VOC12)  \\
         \hline
         two-stage detectors & & &  &  & \\
         \hline
         Faster R-CNN & VGG-16 & 1000$\times$600 & 7 & 73.2 &70.4 \\
         HyperNet & VGG-16 & 1000$\times$600 & 0.88 & 76.3 &71.4\\
         DeRPN & VGG-16 & 1000$\times$600 & - & 76.5 &71.9 \\
         Faster R-CNN & ResNet-101 &1000$\times$600 & 2.4 & 76.4 &73.8\\
         R-FCN & ResNet-101 &1000$\times$600 & 9 & 80.5 &77.6\\
         CoupleNet& ResNet-101 &1000$\times$600 & 8.2 & 82.7 &80.4\\
         
         \hline
         one-stage detectors & & &  &  & \\
         \hline
         YOLO & GoogleNet& 448$\times$448 &45 & 63.4 & 57.9\\
         SSD300 & VGG-16 & 300$\times$300 & 46 & 74.3 & 72.4\\
         SSD512 & VGG-16 & 512$\times$512 & 19 & 76.8 & 74.9\\
         YOLOv2& Darknet-19 & 544$\times$544 &40 &78.6 &  73.4\\
         DSSD321& ResNet-101 & 321$\times$321 &9.5 &78.6  &76.3 \\
         DSSD513& ResNet-101 & 513$\times$513 &5.5 &81.5 & 80.0\\
         YOLOv3 & Darknet-53 &416$\times$416 & 47 &80.25 & -\\
         RefineDet512 &VGG-16& 512$\times$512 & 24.1 &81.8& 80.1\\
         DFPR-Net512&VGG16&512$\times$512 & -&81.1&80.0\\
         DR-Net512& ResNet-101&512$\times$512 & -&82.0&80.4\\
         \hline
         ours & & &  &  & \\
          \hline
         Dubox320 & VGG-16 &320$\times$320 &\textbf{50}&79.31 & 78.82\\
         Dubox512 & VGG-16 &512$\times$512 &18&81.02 & 80.36\\
         Dubox800 & VGG-16 &800$\times$800 &7& 82.31 &81.75\\
         Dubox800(multi-scale) & VGG-16 &800$\times$800 &-& \textbf{82.89} & \textbf{82.01}\\
         \hline
         
    \end{tabular}
    \label{tab:vocperformance}
\end{table*}

\subsubsection{No-prior box detection}

 We also experiment how the fixed hooks in classification and regression affect the detection performance. In Tab.\ref{table:ablationexpriment}, it is clear that using fixed hooks to regress $\left( \varDelta w_1,\varDelta w_2,\varDelta h_1,\varDelta h_2 \right)$  outperforms methods which predict $(c_{x},c_{y},w,h)$  by \textbf{0.8\%} mAP. This also confirms that the hook-based approach has better performance than traditional methods.

\begin{table*}[h]
    \centering
    \caption{Comparison with state-of-the-art detectors on MS COCO test-dev}
    \begin{tabular}{|c|c|c|cccccc|}
        \hline
         Methods & Backbone & Data & AP & AP\footnotesize50 & AP\footnotesize75 & AP\footnotesize S & AP\footnotesize M & AP\footnotesize L  \\
         \hline
         two-stage detectors  &  & & &&&&&\\
         \hline
         Faster R-CNN & VGG-16 & trainval  & 21.9 & 42.7 & - &- &-&- \\
         DeRPN & VGG-16 & trainval  & 25.5& 47.3 &25.4 &9.2 &26.9& 38.3 \\
         R-FCN & ResNet-101 & trainval & 29.9 &51.9& - & 10.8 & 32.8 & 45.0 \\
         CoupleNet& ResNet-101 & trainval & 34.4 &54.8 & 37.2 &13.4&38.1&50.8\\
         
         Deformable R-FCN & Aligned-Inception-ResNet & trainval & 37.5 &58.0 &40.8 &19.4 &40.1 &52.5\\
         \hline
         one-stage detectors&  & & &&&&& \\
         \hline
         YOLOv2& Darknet-19 &trainval35k & 21.6 &44.0 &19.2 &5.0 &22.4 &35.5 \\
         DSSD321& ResNet-101 &trainval35k & 28.0 &46.1 &29.2 &7.4 &28.1 &47.6 \\
         RFB-Net300& VGG-16 & trainval35k &30.3 &49.3 &31.8& 11.8& 31.9 &45.9\\
         DSSD513& ResNet-101 &trainval35k & 33.2 &53.3 &35.2 &13.0 &35.4 &51.1\\
         YOLOv3 608& Darknet-53 &trainval35k & 33.0 &57.9 &34.4 &18.3 &35.4 &41.9\\
         RFB-Net512 &VGG-16 &trainval35k & 33.8 &54.2 &35.9& 16.2& 37.1& 47.4\\
         RetinaNet500& ResNet-101&trainval35k & 34.4 &53.1 &36.8 &14.7 &38.5 &49.1\\
         DFPR-Net512 & ResNet-101 &  trainval35k & 34.6& 54.3& 37.3& 14.7& 38.1 &51.9\\
         RefineDet512 & ResNet-101&trainval35k &36.4& 57.5& 39.5& 16.6& 39.9 &51.4\\
         DR-Net512 & ResNet-101&trainval35k &39.3& 59.8& -& 21.7& 43.7 &50.9\\
         \hline
         ours  &  & & &&&&&\\
          \hline
         Dubox320 & ResNet-101 &trainval35k &31.8 &53.36 &34.01 &17.79 &35.89 &42.87\\
         Dubox512 & ResNet-101 &trainval35k &35.32 &54.75 &37.63 &18.94 &38.80 &45.65\\
         Dubox800 & ResNet-101 &trainval35k &38.03 &56.31 &41.7 &19.01 &40.60 &49.23\\
         Dubox800(multi-scale) & ResNet-101 &trainval35k &\textbf{39.52} &\textbf{57.31} &\textbf{42.18} &\textbf{20.94} &\textbf{41.80} &\textbf{50.86}\\
         \hline
         
    \end{tabular}
    \label{tab:cocoperformance}
\end{table*}
\subsubsection{Importance of residual unit and why dual scales}
We construct a baseline by cutting the box bridge and refine module in Dubox to isolate the direct communication between the detectors, 
and this proves that these two components have a performance improvement of \textbf{2.81\%} mAP, as shown in Tab.\ref{table:ablationexpriment}. 

We also find that $detector_2$ has to be trained more iterations than $detector_1$ for achieving the optimal detection performance. Specifically, the loss of $detector_1$ is stable after $80k$ training iterations, but the mAP of $detector_2$ continues to increase until $140k$ iterations. We suppose that the optimization goal of $detector_2$ is based on the residual of $detector_1$. The convergence of $detector_2$ needs to be under the premise of $detector_1$. So the time to convergence for $detector_2$ is longer than $detector_1$. Under the limits of training time, we only trained Dubox with dual scale detectors which already achieves high detection performance, but we hypothesis more detectors may further improve the performance with our residual unit which can be studied in further work.

\subsubsection{Importance of CRPS loss?}
To demonstrate the effectiveness of the proposed CRPS loss, we remove the IoU gate unit between the classification loss and regression loss. In this setting, no ground truth box is filtered, and all ground truth boxes are used in computing loss for both detectors. Hooks corresponding to positive sample compute classification and regression loss, and others corresponding to negative samples only compute in classification loss. Tab.\ref{table:ablationexpriment} shows that this setting leads to drop \textbf{2.42\%} mAP compared to the method with IoU gate unit. This significant performance decline indicates that the CRPS loss is effective for Dubox.

\subsubsection{Importance of reducing redundancy}
We also compare the detection performance of dual scale detector on different object categories with the reducing redundancy strategy. As shown in Tab.\ref{tab:dualshotperformace}, we find that the mAP on small object, such as bottle, of $detector_1$ is always higher than  $detector_2$, and the mAP on large object, such as airplane, of $detector_2$ is higher than $detector_1$. The performance gaps of the detectors is further enlarged by the mechanism of reducing redundancy strategy. This result indicates that the reducing redundancy strategy helps the two detectors to focus on detecting different scales objects.
\subsection{Comparisons with state-of-the-art methods}
\subsubsection{VOC}
VOC 2007 $trainval$, $test$ and VOC 2012 $trainval$ set are used as the training data. We set the batch size to 50(320$\times$320), 32(512$\times512$), 16(800$\times$800) for each GPU in training, and train the model with $10^{-3}$ learning rate for the first 80k iterations, then reduce to $10^{-4}$ for another $40k$ iterations, respectively. We evaluate our detector on VOC07 and VOC12 $test$ set. As the performance show in Tab.\ref{tab:vocperformance}, the proposed Dubox detector gets \textbf{81.02\%} on VOC07 and \textbf{80.36\%} on VOC12 with input size 512 $\times$ 512, and get \textbf{82.31\%} and \textbf{81.75\%} when the input size is 800$\times$800.

Even feeding with 320$\times$320 input size, Dubox obtains the top \textbf{79.31\%}mAP on VOC07 and \textbf{78.82\%} on VOC12, which is even better than most of those two-stage methods using about 1000 $\times$ 600 input size (e.g., 70.4\% of Faster R-CNN \cite{fasterrcnn}, 76.5\% of DeRPN\cite{derpn} and 77.6\% of R-FCN \cite{rfcn}). With the input size 512 $\times$ 512 or 800 $\times$ 800, the performance is surpassing the one-stage methods(e.g., 63.4, 57.9  in YOLO\cite{yolov1}, 78.6, 73.4 in YOLOv2\cite{yolov2}, 80.25 in YOLOv3 \cite{yolov3voc}, 78.6, 76.3 in DSSD\cite{dssd},81.8, 80.1 in RefineDet\cite{refinedet}, 81.2\% in DFPR-Net\cite{dfprnet}, 82.0\% in DR-Net\cite{drnet}). Meanwhile, we experiment Dubox800 with multi-scale images, it gets \textbf{82.89}, \textbf{82.01\%}mAP.

\subsubsection{COCO}

We also evaluate Dubox on MS COCO\cite{mscoco}. Unlike PASCAL VOC, we  report the results of ResNet-101 based DuBox directly. Following the protocol in MS COCO, $trainval35k$ set \cite{mscoco} is used for training and evaluate the results on $test$-$dev$ set. We set the batch size to 16(320$\times$320), 8(512$\times512$), 4(800$\times$800) for each GPU in training, and train the model with $10^{-3}$ learning rate for the first 280k iterations, then $10^{-4}$ and $10^{-5}$ for another $180k$ and $140k$ iterations, respectively.

As show in Tab.\ref{tab:cocoperformance} on MS COCO test set. Dubox320 with ResNet-101 produces \textbf{31.8\%} AP that is better than  other two-stage methods (e.g., Faster R-CNN\cite{fasterrcnn}, DeRPN\cite{derpn}). The accuracy of Dubox can be improved to \textbf{35.32\%} using 512$\times$512 input size. When the input size is 800 $\times$ 800, Dubox gets \textbf{38.03\%}AP  which is much better than several one-stage object detectors (e.g., SSD\cite{ssd} and YOLOv2\cite{yolov2}, YOLOv3\cite{yolov3coco}, RetinaNet\cite{retinanet}, RFB-Net\cite{rfbnet}, DFPR-Net\cite{dfprnet}, DR-Net\cite{drnet}.). With multi-scale input, Dubox800 get \textbf{39.52\%}AP.

\subsection{Inference time Performance}
With the help of no-prior box detection, Dubox only uses 4 channel feature maps to represent the target boxes, while the anchor-based methods have to use $4A$ ($A$ is the number of anchors) feature maps to regress the boxes. The less output feature maps and less scale detectors accelerate the inference speed of Dubox. Furthermore, the architecture of proposed Dubox only contains convolution, deconvolution layer and element-wise activation functions which are highly optimized in common deep learning frameworks, making it easy to deploy in resource limited platforms, such as mobile phone and autopliot systems.

We present the inference speed of Dubox and the state-of-the-art methods in the Tab.\ref{tab:vocperformance}. The speed is evaluated with batch size 1 on a machine with NVIDIA Titan X, CUDA 8.0 and cuDNN v6. As shown in Tab.\ref{tab:vocperformance}, we find that Dubox processes an image in \textbf{20.0}ms (\textbf{50} FPS), \textbf{55.4}ms (\textbf{18} FPS) and \textbf{142}ms (\textbf{7} FPS) with input sizes 320$\times$320, 512$\times$512 and 800$\times$800 respectively. To the best of our knowledge, Dubox is the first getting \textbf{50} FPS real-time method to achieve detection accuracy above \textbf{79.31\%} mAP on PASCAL VOC 2007. In summary, Dubox achieves the best trade-off between accuracy and speed.
\section{Conclusion}
Anchor-based method is not the only choice in object detection. Dubox, as an no-prior box method,  also can work effectively using the proper regression loss and network architecture. At the same time, Dubox further considered the problem of multi-scale detection to enhance the capacity of heuristic-guided feature selection. The proposed dual scale residual unit enables multi-scale detectors not to operate independently, but make the high-level learning from the low one. These strategies improve the performance of the detection while significantly reducing the redundancy of various scale detectors.

{\small
\bibliographystyle{ieee}
\bibliography{dubox}

\begin{thebibliography}{10}\itemsep=-1pt

\bibitem{rfcn}
J.~Dai, Y.~Li, K.~He, and J.~Sun.
\newblock R-fcn: Object detection via region-based fully convolutional
  networks.
\newblock In {\em Advances in neural information processing systems}, pages
  379--387, 2016.

\bibitem{hog}
N.~Dalal and B.~Triggs.
\newblock Histograms of oriented gradients for human detection.
\newblock In {\em international Conference on computer vision \& Pattern
  Recognition (CVPR'05)}, volume~1, pages 886--893. IEEE Computer Society,
  2005.

\bibitem{voc}
M.~Everingham, L.~Van~Gool, C.~K.~I. Williams, J.~Winn, and A.~Zisserman.
\newblock The pascal visual object classes (voc) challenge.
\newblock {\em International Journal of Computer Vision}, 88(2):303--338, June
  2010.

\bibitem{dssd}
C.-Y. Fu, W.~Liu, A.~Ranga, A.~Tyagi, and A.~C. Berg.
\newblock Dssd: Deconvolutional single shot detector.
\newblock {\em arXiv preprint arXiv:1701.06659}, 2017.

\bibitem{xavier}
X.~Glorot and Y.~Bengio.
\newblock Understanding the difficulty of training deep feedforward neural
  networks.
\newblock In {\em Proceedings of the thirteenth international conference on
  artificial intelligence and statistics}, pages 249--256, 2010.

\bibitem{residualnet}
K.~He, X.~Zhang, S.~Ren, and J.~Sun.
\newblock Deep residual learning for image recognition.
\newblock In {\em Proceedings of the IEEE conference on computer vision and
  pattern recognition}, pages 770--778, 2016.

\bibitem{densebox}
L.~Huang, Y.~Yang, Y.~Deng, and Y.~Yu.
\newblock Densebox: Unifying landmark localization with end to end object
  detection.
\newblock {\em arXiv preprint arXiv:1509.04874}, 2015.

\bibitem{dfprnet}
T.~Kong, F.~Sun, C.~Tan, H.~Liu, and W.~Huang.
\newblock Deep feature pyramid reconfiguration for object detection.
\newblock In {\em Proceedings of the European Conference on Computer Vision
  (ECCV)}, pages 169--185, 2018.

\bibitem{cornernet}
H.~Law and J.~Deng.
\newblock Cornernet: Detecting objects as paired keypoints.
\newblock In {\em Proceedings of the European Conference on Computer Vision
  (ECCV)}, pages 734--750, 2018.

\bibitem{li2015convolutional}
H.~Li, Z.~Lin, X.~Shen, J.~Brandt, and G.~Hua.
\newblock A convolutional neural network cascade for face detection.
\newblock In {\em Proceedings of the IEEE conference on computer vision and
  pattern recognition}, pages 5325--5334, 2015.

\bibitem{liang2018cirl}
X.~Liang, T.~Wang, L.~Yang, and E.~Xing.
\newblock Cirl: Controllable imitative reinforcement learning for vision-based
  self-driving.
\newblock In {\em Proceedings of the European Conference on Computer Vision
  (ECCV)}, pages 584--599, 2018.

\bibitem{fpn}
T.-Y. Lin, P.~Doll{\'a}r, R.~Girshick, K.~He, B.~Hariharan, and S.~Belongie.
\newblock Feature pyramid networks for object detection.
\newblock In {\em Proceedings of the IEEE Conference on Computer Vision and
  Pattern Recognition}, pages 2117--2125, 2017.

\bibitem{retinanet}
T.-Y. Lin, P.~Goyal, R.~Girshick, K.~He, and P.~Doll{\'a}r.
\newblock Focal loss for dense object detection.
\newblock In {\em Proceedings of the IEEE international conference on computer
  vision}, pages 2980--2988, 2017.

\bibitem{mscoco}
T.-Y. Lin, M.~Maire, S.~Belongie, J.~Hays, P.~Perona, D.~Ramanan,
  P.~Doll{\'a}r, and C.~L. Zitnick.
\newblock Microsoft coco: Common objects in context.
\newblock In {\em European conference on computer vision}, pages 740--755.
  Springer, 2014.

\bibitem{rfbnet}
S.~Liu, D.~Huang, et~al.
\newblock Receptive field block net for accurate and fast object detection.
\newblock In {\em Proceedings of the European Conference on Computer Vision
  (ECCV)}, pages 385--400, 2018.

\bibitem{ssd}
W.~Liu, D.~Anguelov, D.~Erhan, C.~Szegedy, S.~Reed, C.-Y. Fu, and A.~C. Berg.
\newblock Ssd: Single shot multibox detector.
\newblock In {\em European conference on computer vision}, pages 21--37.
  Springer, 2016.

\bibitem{sift}
D.~G. Lowe.
\newblock Object recognition from local scale-invariant features.
\newblock In {\em iccv}, page 1150. Ieee, 1999.

\bibitem{yolov1}
J.~Redmon, S.~Divvala, R.~Girshick, and A.~Farhadi.
\newblock You only look once: Unified, real-time object detection.
\newblock In {\em Proceedings of the IEEE conference on computer vision and
  pattern recognition}, pages 779--788, 2016.

\bibitem{yolov2}
J.~Redmon and A.~Farhadi.
\newblock Yolo9000: better, faster, stronger.
\newblock In {\em Proceedings of the IEEE conference on computer vision and
  pattern recognition}, pages 7263--7271, 2017.

\bibitem{yolov3coco}
J.~Redmon and A.~Farhadi.
\newblock Yolov3: An incremental improvement.
\newblock {\em arXiv preprint arXiv:1804.02767}, 2018.

\bibitem{fasterrcnn}
S.~Ren, K.~He, R.~Girshick, and J.~Sun.
\newblock Faster r-cnn: Towards real-time object detection with region proposal
  networks.
\newblock In {\em Advances in neural information processing systems}, pages
  91--99, 2015.

\bibitem{imagenet}
O.~Russakovsky, J.~Deng, H.~Su, J.~Krause, S.~Satheesh, S.~Ma, Z.~Huang,
  A.~Karpathy, A.~Khosla, M.~Bernstein, et~al.
\newblock Imagenet large scale visual recognition challenge.
\newblock {\em International journal of computer vision}, 115(3):211--252,
  2015.

\bibitem{ohem}
A.~Shrivastava, A.~Gupta, and R.~Girshick.
\newblock Training region-based object detectors with online hard example
  mining.
\newblock In {\em Proceedings of the IEEE Conference on Computer Vision and
  Pattern Recognition}, pages 761--769, 2016.

\bibitem{vgg16}
K.~Simonyan and A.~Zisserman.
\newblock Very deep convolutional networks for large-scale image recognition.
\newblock {\em arXiv preprint arXiv:1409.1556}, 2014.

\bibitem{tang2018pyramidbox}
X.~Tang, D.~K. Du, Z.~He, and J.~Liu.
\newblock Pyramidbox: A context-assisted single shot face detector.
\newblock In {\em Proceedings of the European Conference on Computer Vision
  (ECCV)}, pages 797--813, 2018.

\bibitem{iouloss}
J.~Wang, Y.~Yuan, G.~Yu, and S.~Jian.
\newblock Sface: An efficient network for face detection in large scale
  variations.
\newblock {\em arXiv preprint arXiv:1804.06559}, 2018.

\bibitem{attention}
S.~Woo, J.~Park, J.-Y. Lee, and I.~So~Kweon.
\newblock Cbam: Convolutional block attention module.
\newblock In {\em The European Conference on Computer Vision (ECCV)}, September
  2018.

\bibitem{derpn}
L.~Xie, Y.~Liu, L.~Jin, and Z.~Xie.
\newblock Derpn: Taking a further step toward more general object detection.
\newblock {\em arXiv preprint arXiv:1811.06700}, 2018.

\bibitem{drnet}
H.~Xu, X.~Lv, X.~Wang, Z.~Ren, N.~Bodla, and R.~Chellappa.
\newblock Deep regionlets for object detection.
\newblock In {\em The European Conference on Computer Vision (ECCV)}, September
  2018.

\bibitem{refinedet}
S.~Zhang, L.~Wen, X.~Bian, Z.~Lei, and S.~Z. Li.
\newblock Single-shot refinement neural network for object detection.
\newblock In {\em Proceedings of the IEEE Conference on Computer Vision and
  Pattern Recognition}, pages 4203--4212, 2018.

\bibitem{yolov3voc}
Z.~Zhang, T.~He, H.~Zhang, Z.~Zhang, J.~Xie, and M.~Li.
\newblock Bag of freebies for training object detection neural networks.
\newblock {\em arXiv preprint arXiv:1902.04103}, 2019.

\end{thebibliography}
}

\end{document}